\documentclass[10pt,twocolumn,letterpaper]{article}

\usepackage{wacv}
\usepackage{times}
\usepackage{epsfig}
\usepackage{graphicx}
\usepackage{amsmath}
\usepackage{amssymb}
\usepackage{color}
\usepackage{import}
\usepackage{preamble}
\usepackage{wrapfig}
\usepackage[utf8]{inputenc}
\usepackage[english]{babel}
\usepackage[numbers]{natbib}
\usepackage{array}
\usepackage{lipsum}
\usepackage{hyperref}

\wacvfinalcopy 

\def\wacvPaperID{535} 

\ifwacvfinal\fi
\begin{document}

\title{TextCaps : Handwritten Character Recognition with Very Small Datasets}

\author{Vinoj Jayasundara \\
University of Moratuwa\\
{\tt\small vinojjayasundara@gmail.com}
\and
Sandaru Jayasekara \\
University of Moratuwa\\
{\tt\small sandaruamashan@gmail.com}
\and
Hirunima Jayasekara \\
University of Moratuwa\\
{\tt\small nhirunima@gmail.com}
\and
Jathushan Rajasegaran \\
University of Moratuwa\\
{\tt\small brjathu@gmail.com}
\and
Suranga Seneviratne \\
University of Sydney\\
{\tt\small suranga.seneviratne@sydney.edu.au}
\and
Ranga Rodrigo \\
University of Moratuwa\\
{\tt\small ranga@uom.lk}
}

\maketitle
\ifwacvfinal\thispagestyle{empty}\fi

\begin{abstract}
Many localized languages struggle to reap the benefits of recent advancements in character recognition systems due to the lack of substantial amount of labeled training data. This is due to the  difficulty in generating large amounts of labeled data for such languages and inability of deep learning techniques to properly learn from small number of training samples. We solve this problem by introducing a technique of generating new training samples from the existing samples, with realistic augmentations which reflect actual variations that are present in human hand writing, by adding random controlled noise to their corresponding instantiation parameters. Our results with a mere 200 training samples per class surpass existing character recognition results in the EMNIST-letter dataset while achieving the existing results in the three datasets: EMNIST-balanced, EMNIST-digits, and MNIST. We also develop a strategy to effectively use a combination of loss functions to improve reconstructions. 
Our system is useful in character recognition for localized languages that lack much labeled training data and even in other related more general contexts such as object recognition. 
\footnote{\textcolor{blue}{\url{https://github.com/vinojjayasundara/textcaps}}}

\end{abstract}

\section{Introduction}
\label{sect:intro}



Handwritten character recognition is a nearly solved problem for many of the mainstream languages thanks to the recent advancements in deep learning models \cite{cirecsan2010deep}. Nonetheless, for many other languages, handwritten digit recognition remains a challenging problem due to the lack of sufficiently large labeled datasets that are essential to train deep learning models \cite{Krizhevsky:2012:ICD:2999134.2999257}. While conventional models such as linear classifiers, K-nearest neighbors, non-linear classifiers, and Support Vector Machines (SVM)  \cite{lee1991handwritten} can be used for this task, they are not able to achieve the near human level performances provided by deep learning models. Convolutional Neural Networks (CNN) have achieved state-of-the-art results due to their ability to encode deep features and spatial understanding. Although CNNs are good at understanding low level and high level features in images, by doing so, they lose valuable information at pooling layers. CNNs require large number of training samples (usually in the scale of thousands or tens of thousands per class) to train and classify images successfully. As a result, there is a strong interest in training CNNs with a lesser number of training samples.

In this paper, we propose a technique which tackles this problem of the labeled dataset being small in size, with the aid of Capsule Networks (CapsNets) \cite{sabour2017dynamic}. We exploit their ability to augment data just by manipulating the instantiation parameters \cite{hinton2011transforming}. CapsNets learn the properties of an image---in this case a character---in addition to its existence. This makes them useful in learning to recognize characters with a less amount of labeled data. Our architecture is based on the CapsNet architecture proposed by Sabour \textit{et al.} \cite{sabour2017dynamic}, which comprises a capsule network and a fully connected decoder network. We replace the decoder network with a deconvolutional network while doing minor alterations to the capsule network. By adding a controlled amount of noise to the instantiation parameters that represent the properties of an entity, we transform the entity to characterize actual variations that happen in reality.  This results in a novel data generation technique, much more realistic than augmenting data with affine transformations. As the reconstruction accuracy is also important in many contexts, we present an empirically appropriate strategy of combining loss functions which significantly improves the reconstruction. 
Our system achieves results that are on-par with the state-of-the-art with just 200 data points per class, while achieving even better results with larger volumes of training data. 




The key contributions of this paper are as follows:

\begin{itemize}[noitemsep,nolistsep]
\item  We outperform the state-of-the-art results in EMNIST-letters, EMNIST-balanced and EMNIST-digits character datasets, by training our
system on all the training samples available.

\item We evaluate the proposed architecture on a non-character dataset, Fashion-MNIST, to ensure the
flexibility and robustness.  We achieve very good results with 200 training samples and achieve the state-of-the-art with the full dataset.

\item We propose a novel technique for training capsule networks with small number
of training samples, as small as 200 per class, and keeping the same set of
test samples, while achieving the state-of-the-art performance. Our method
require only 10\% of data necessary for a state-of-the-art system, to produce similar results.

\item  We propose and evaluate several variations to the decoder network and analyze
its performance with different loss functions to provide a strategy to
select a suitable combination of loss functions.

\end{itemize}

Rest of the paper is organized as follows: in Section \ref{sect:related} we discuss the related works, and in Section \ref{sec:meth_overview} we explain our methodology. Subsequently, we discuss our results in Section \ref{sec:results}.

\section{Related Work}
\label{sect:related}
MNIST \cite{lecun1998mnist} is the widely used benchmark for the handwritten digit recognition task. Multiple works \cite{article,DBLP:journals/corr/abs-1202-2745,4668644,Ranzato:2006:ELS:2976456.2976599,5459469} have used CNN models on MNIST dataset and have achieved results in excess of 99\% accuracy. 
Apart from digit recognition, several attempts \cite{2017arXiv170205373C,2017arXiv170909161D} have been made in handwritten character recognition with EMNIST datasets \cite{2017arXiv170205373C}.
A bidirectional neural network is introduced in \cite{DBLP:journals/corr/abs-1803-01900} which is capable of performing both image classification and image reconstruction by adding a style memory to the output layer of the network. 

The idea of a capsule was introduced in 2011 by \cite{hinton2011transforming}, as a transforming autoencoder. With a three layered CapsNets architecture, and by training the network using dynamic routing, authors of \cite{sabour2017dynamic} have achieved 0.25\% error rate on the MNIST dataset. This architecture consists of a primary capsule, which was built by stacking convolutional layers, and a fully connected capsule, which uses routing by agreement between higher level capsules and lower level capsules. We draw intuition for this paper from the concept of instantiation parameters proposed in \cite{hinton2011transforming}, while emphasizing that our work is significantly novel and different from \cite{hinton2011transforming}.



We identified two main solutions in the literature to the low data issue, namely one-shot learning and new data generation. An example of the former is the Siamese networks as proposed in \cite{Koch2015SiameseNN}. A one-shot learning deep model was proposed by Bertinetto \textit{et al.}\cite{DBLP:journals/corr/BertinettoHVTV16}, where they used a second network to predict the parameters of the first network. Since one-shot learning solutions are mostly application-specific, we turn to a new data generation approach.

Existing literature offers several successful data generation techniques. Although GANs \cite{goodfellow2014generative} can be used to generate data, a basic form of a GAN network will not be sufficient to augment the dataset for training, since it can not generate labelled data, unless a separate GAN is trained per class.
Another potential choice which has image generation capabilities, Variational Autoencoders (VAEs) \cite{doersch2016tutorial} have similar problems. VAEs represent all the images as 1D vectors,
whereas capsule networks have dedicated dimensions for each class. As a result, when VAE’s 1D vectors are perturbed, there is a high probability that those changes affect multiple classes. 
Data augmentation techniques such as jittering and flipping (not
suitable for characters) offer limited amount of simple augmentation. Thus, they can not offer complex and subtle variations that are more closer to the human variations. A comprehensive comparison of our results with these techniques are provided in Section \ref{sec:results_perturb}.


\section{Methodology} \label{sec:meth_overview}

This sections outlines our approach. Prior to experimenting with reduced datasets, in Section \ref{sec:meth_full}, we attempt to surpass the state-of-the-art results for several hand written digit databases including EMNIST balanced, EMNIST letters and EMNIST digits, with the use of all the training samples provided. Subsequently, we attempt to achieve the state-of-the-art performance using a limited number of training samples, as low as 200 training samples per class, as opposed to, for example, 2400 data points per class in EMNIST balanced and 4800 data points per class in EMNIST letters.

In order to address the drawbacks faced when training the classifier with low number of training samples, we propose a novel technique for increasing the number of training samples in Section \ref{sec:meth_perturb}. We perform a comprehensive analysis of the effect of the loss function in reconstruction, in Section \ref{sec:meth_loss}.
\subsection{Character Recognition with Capsule Networks} 
\label{sec:meth_full}

\begin{figure*}[!h]
  \centering
  \includegraphics[scale=0.5]{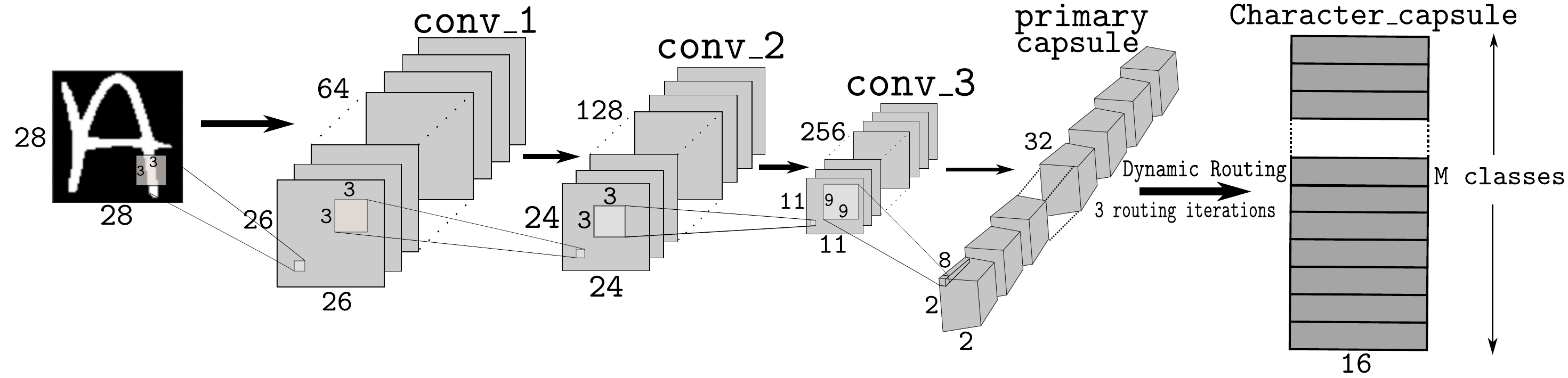}
  \vspace{-2mm}
  \caption{\textbf{TextCap Model}: Proposed CapsNet model for character classification.}
 \label{fig:base}
 \vspace{-3mm}
\end{figure*}

\begin{figure*}[!h]
  \centering
  \includegraphics[scale=0.2]{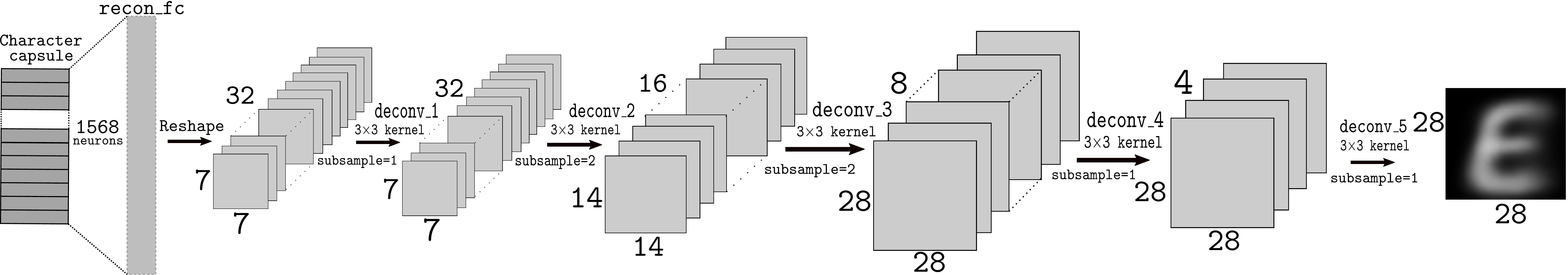}
  \caption{\textbf{TextCap Decoder}: Decoder network for the character reconstruction. Input to this network is obtained by masking the DigitCaps layer of the TextCap classifier}
 \label{fig:decoder}
  \vspace{-4mm}
\end{figure*}

For the character recognition task, we propose an architecture comprising of a capsule network and a decoder network, as illustrated in Fig. \ref{fig:base} and Fig. \ref{fig:decoder}. 

In the capsule network, the first three layers are convolutional layers with 64 3$\times$3 kernels with stride 1, 128 3$\times$3 kernels with stride 1 and 256 3$\times$3 kernels with stride 2 respectively. The fourth layer is a primary capsule layer with 32 channels of 8-dimensional capsules, with each primary capsule containing 8 convolutional units with a 9$\times$9 kernel and a stride of 2. The fifth layer, termed as the character capsule layer, is a fully connected capsule layer with a 16-dimensional capsule per class, resulting in $M$ capsules for a dataset with $M$ number of classes. We use dynamic routing between the primary capsule layer and the character capsule layer, as proposed by \cite{sabour2017dynamic}, with 3 routing iterations. The input to the capsule network is a set of $J$, $28\times28$ images and the output is a $J\times M \times16$ dimensional tensor $C$, containing the corresponding instantiation parameters, where each $C_{j}, j\in[J]$ is the instantiation parameter matrix of the $j^{th}$ training sample.



Prior to passing $C$ as the input to the decoder network, the corresponding instantiation parameters should be masked with zeros for all the classes except the true class. Hence, the masked tensor $\widehat{C}$ is still a $J \times M \times16$ dimensional matrix, yet containing only the instantiation parameters corresponding to the true class as the non-zero values.

The decoder network comprises one fully connected layer, followed by five deconvolutional layers \cite{decon2010} with parameters as shown by Fig. \ref{fig:decoder}. The input to the decoder is the masked matrix $\widehat{C}$, and the output of the decoder is the set of reconstructed $28\times28$ images. Except for the final deconvolution layer, which has sigmoid activation, the fully connected layer and the other deconvolution layers have ReLU activation.

First, we train the proposed model with the full training sets and evaluate its performance. Subsequently, in an effort to address the issue of lack of high number of training samples in character recognition and similar tasks as elaborated in section \ref{sect:intro}, we attempt to achieve the state-of-the-art performance using 200 training samples per class, using the same network.


By examining the results of the above section with low number of training samples, as elaborated in Section \ref{sec:results_classification}, we observed that even though the capsule network performance achieved the state-of-the-art, the decoder network fails to achieve acceptable reconstruction. The most obvious solution to enhancing the performance of the decoder network is to increase the number of training samples, by generating new training samples from the samples available in the original (reduced) training set. Therefore, we propose a novel method of generating new training samples by augmenting original training samples with the aid of the concept of instantiation parameters in the CapsNets, as described by the following Section \ref{sec:meth_perturb}

\subsection{Proposed Technique for Image Data Generation Using Perturbation of Instantiation Parameters}
\label{sec:meth_perturb}

\begin{figure}[!h]
 \vspace{-2mm}
  \centering
  \includegraphics[scale=0.2]{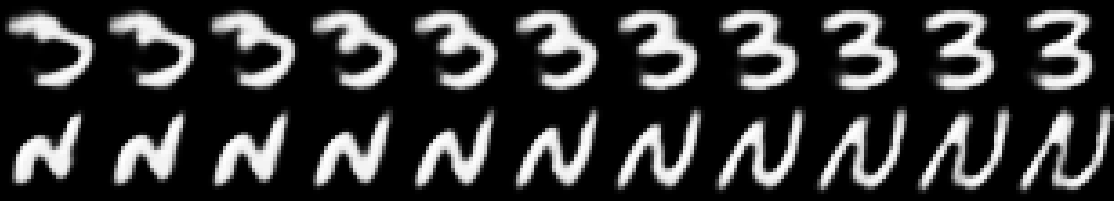}
  \caption{Variation in characters with the perturbation of instantiation parameters}
 \label{fig:inst_change}
 \vspace{-2mm}
\end{figure}

From the concept of instantiation parameters in capsule networks, we can represent any character using a 16 dimensional vector \cite{sabour2017dynamic}. With a pre-trained decoder network, we can successfully reconstruct the original image, by using only this instantiation parameter vector. The intuition behind our proposed perturbation algorithm is that by adding controlled random noise to the values of the instantiation vector, we can create new images, which are significantly different from the original images, effectively increasing the size of the training dataset. Fig. \ref{fig:inst_change} illustrates the variations of an image, when one particular instantiation parameter is changed thusly. 

\begin{figure*} [h]
 \centering
    \subfigure[Initial Training of $M_1$ \label{fig:sys_1}]
    {\includegraphics[scale=0.45,clip]{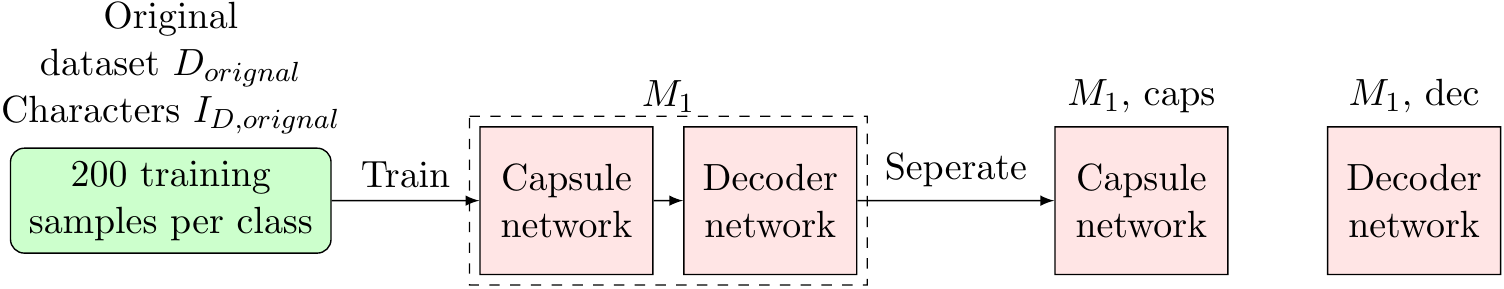}}
    \hspace{1.1cm}
    \subfigure[Generating instantiation parameters and reconstructed images \label{fig:sys_2}]
    {\includegraphics[scale=0.45]{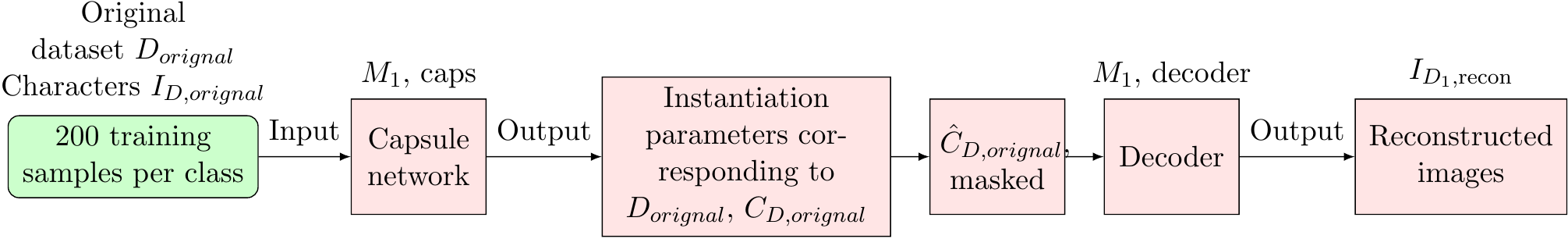}}
    \subfigure[Proposed decoder re-training technique \label{fig:sys_3}]
    {\includegraphics[scale=0.45]{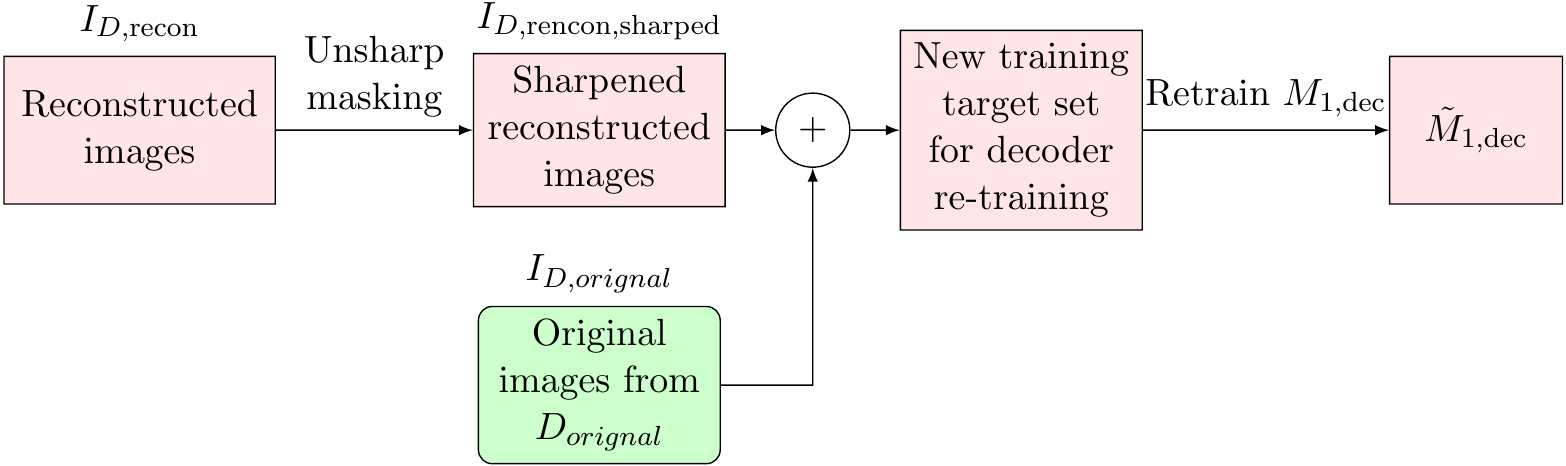}}
    \hspace{1cm}
    \subfigure[New image data generation by the proposed technique \label{fig:sys_4}]
    {\includegraphics[scale=0.5]{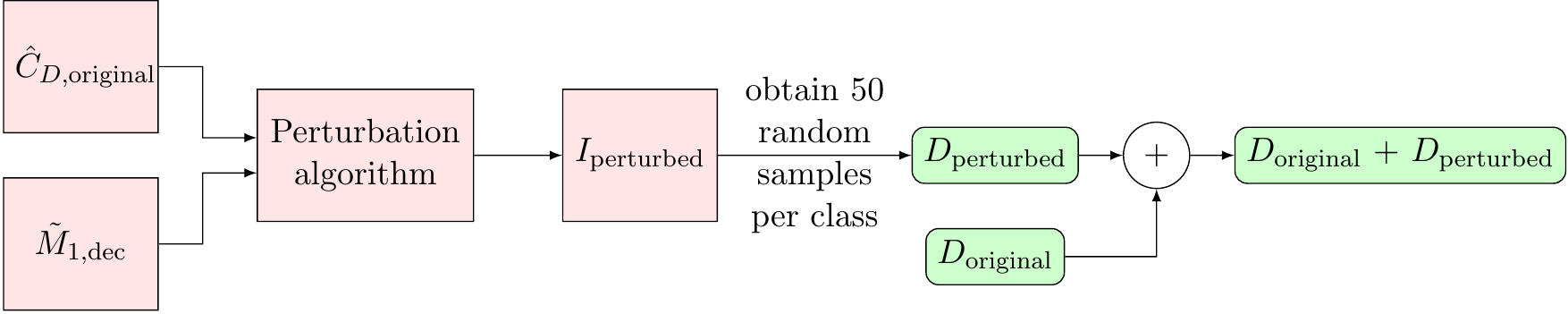}}
    \subfigure[Training model $M_2$ afresh with the new dataset\label{fig:sys_5}]
    {\includegraphics[scale=0.45]{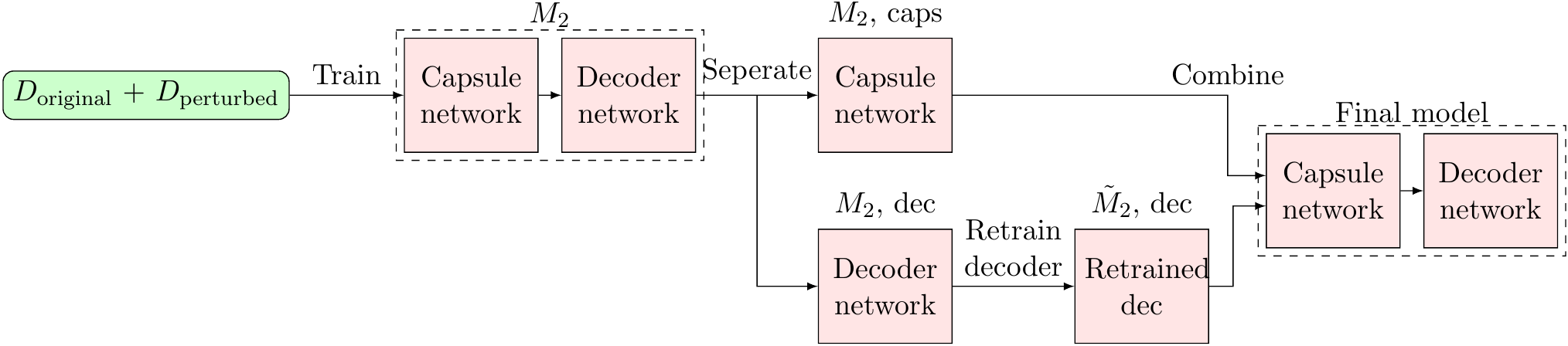}}

\caption{The overall methodology for improving the decoder performance}
\label{fig:system}
\vspace{-4mm}
\end{figure*}

Similarly, each of the instantiation parameter is responsible for a certain property of the image, individually or collectively. Hence, we propose a novel technique of generating a new dataset, from a dataset with limited amount of training samples, as illustrated by Fig. \ref{fig:system} and Algorithm \ref{algo:1}.


\algnewcommand\algorithmicforeach{\textbf{for each}}
\algdef{S}[FOR]{ForEach}[1]{\algorithmicforeach\ #1\ \algorithmicdo}

\begin{figure}[!h]
\vspace{-2mm}
\centering
\subfigure{\includegraphics[scale=0.8]{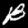}}
\subfigure{\includegraphics[scale=0.8]{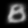}} \hspace{0.5cm}
\subfigure{\includegraphics[scale=0.8]{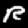}}
\subfigure{\includegraphics[scale=0.8]{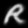}}\hspace{0.5cm}
\subfigure{\includegraphics[scale=0.8]{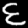}}
\subfigure{\includegraphics[scale=0.8]{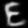}}
\caption{Original and reconstructed Image pairs with Model $M_1$}
\label{fig:poor_recon}
\vspace{-3mm}
\end{figure}

First, as illustrated by Fig. \ref{fig:sys_1}, we train the network proposed in Section \ref{sec:meth_full}, $M_1$, with dataset, $D_{\mathrm{original}}$, containing 200 training samples per class and all testing samples. Without loss of generality, we choose the first 200 training samples in each class in the dataset. Subsequently, we consider the trained capsule network, $M_{1,\mathrm{caps}}$, and the trained decoder network, $M_{1,dec}$, separately. Next, as illustrates by Fig. \ref{fig:sys_2}, we obtain the instantiation parameters, $C_{D,\mathrm{original}}$, corresponding to the training images, $I_{D,\mathrm{original}}$, in $D_{\mathrm{original}}$ as the output of the capsule network $M_{1,\mathrm{caps}}$, which can be masked as $\widehat{C}_{D,\mathrm{original}}$ and passed as the input of the decoder network $M_{1,\mathrm{dec}}$. We can obtain the corresponding reconstructed images $I_{D,\mathrm{recon}}$ as the output of $M_{1,\mathrm{dec}}$. Fig. \ref{fig:poor_recon} shows several such reconstructed images.

From Fig. \ref{fig:poor_recon}, we clearly observe that training with such low number of training samples result in poor reconstruction performance. The subtle variations in the input characters are absent from the reconstructions, in addition to being blurred. Hence, we cannot directly apply the concept of perturbation and create new training samples from a such poorly trained model. First, we attempt to eliminate the blurriness in the decoder network output, by proposing the following technique, illustrated by Fig. \ref{fig:sys_3}. For each reconstructed image in $I_{D,\mathrm{recon}}$, we perform unsharp masking \cite{826787} with $radius=1$, $threshold=1$ and $unsharp$ $strength = 10$ $times$, which sharpens the reconstructed images. Then we combine the new sharpened image set $I_{D,\mathrm{recon,sharped}}$ with the initial $I_{D,\mathrm{recon}}$ set, in order to create a new target set for the decoder $M_{1,\mathrm{dec}}$. Subsequently, we re-train the decoder for 10 epochs with this new target set, in order to obtain an improved decoder $\widetilde{M}_{1,\mathrm{dec}}$ which provides sharper reconstructions than $M_{1,\mathrm{dec}}$.

After training, there can be training samples which are not properly learned, and hence are wrongly reconstructed. If these wrongly reconstructed samples are considered for perturbing and creating new samples, it may result in miss-classified samples in the newly generated training dataset. Therefore, prior to applying perturbation and creating new training samples, it is necessary to remove such training samples, after the model is trained. 

Subsequently, we perform new data generation by perturbation, as illustrated by Fig. \ref{fig:sys_4}. For non-zero instantiation parameters in $\widehat{C}_{D,\mathrm{original}}$, we add random controlled noise (Algorithm \ref{algo:1}). Here, we perturb only one instantiation parameter at a time to generate new samples to avoid distortions. Hence, a method of selection of which instantiation parameter to perturb is required. We observed that, for a given class, there exists a relationship between the variance of an instantiation parameter and the actual physical variations in the generated images. Higher the variance, we observe rapid variations and vice versa. Hence, for each instantiation parameter $k \in [0,15]$ in each class $m \in M$, we calculate the variance, $\sigma_{m,k}$, across all the training samples that belongs to $m$, and sort in the descending order. We have 16 choices for the value of $a$, with $a=0$ representing the instantiation parameter with the highest variance and $a=15$ representing that with lowest variance. For this study, we generate two datasets with $a=0$ and $a=1$.

For a given $a$, we calculate the noise value to add for each instantiation parameter considered. The adjusted value of the instantiation parameter should not exceeded the maximum value that it can take for a given class. Adding noise to instantiation parameters without any restrictions will lead to various distortions in the reconstructed images, as illustrated in Fig. \ref{fig:too_much} below. The \textbf{H} has visually changed classes to \textbf{A}, and \textbf{a} is visually unrecognizable anymore. Such distortions are detrimental for a data generation technique. Therefore, we propose a carefully designed control mechanism to avoid such distortions, without manual elimination by visual inspection. 

\begin{figure}[!h]
\vspace{-2mm}
    \centering
    \includegraphics[scale=0.11]{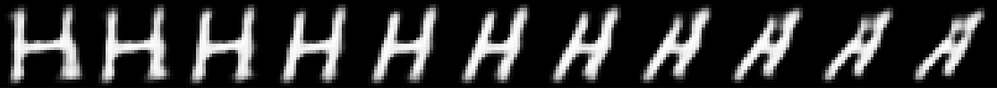}
    \includegraphics[scale=0.11]{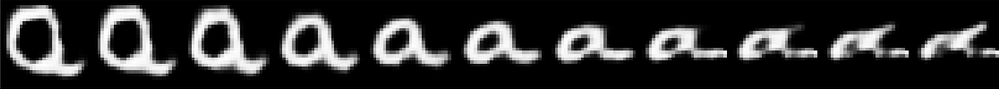}
    \caption{Distortions caused by adding uncontrolled noise}
    \label{fig:too_much}
    \vspace{-3mm}
\end{figure}

Hence, for each instantiation parameter $k$ of the class $m$, we calculate the maximum noise that can be added, $\tau_{m,k}$. We further constrain the noise by $\tau_{k}$, which is the average maximum noise that can be added for the instantiation parameter $k$ across all the classes, even though $\tau_{m,k}$ allows for higher values. This is to prevent sudden high variations occurring after perturbations. Hence, the noise value added for any $k$ is capped at $\tau_{k}$. Subsequently, we obtain the new reconstructed images, $I_{D,\mathrm{perturbed}}$, which are significantly different from the original training images, $I_{D,\mathrm{original}}$, by passing the perturbed instantiation parameter tensor to the decoder $\widetilde{M}_{1,\mathrm{dec}}$.

\renewcommand{\algorithmicrequire}{\textbf{Input:}}
\renewcommand{\algorithmicensure}{\textbf{Output:}}

\begin{algorithm}\caption{Image data generation using perturbation} \label{algo:1}
\algorithmicrequire \ Instantiation parameters $\widehat{C}$, $a^{th}$ highest variance, Decoder Network model ($\widetilde{M}_{dec}$). \newline
\algorithmicensure \ Perturbed images $I_{\mathrm{perturbed}}$

\begin{algorithmic}[1]
\State  Calculate class variance $\sigma_{m,k} = \mathrm{var}_j(\widehat{C}_{m,j,k})$.   
\State  Get $\tilde{\sigma}_{m,k'}$ $\gets$ $sort_k$($\sigma_{m,k}$) descending.    
\State Get $\hat{k}=k$ corresponding to $k' = a$.
\State $ \tau_{m,k} \gets \frac{\max_{j}(\widehat{C}_{m,j,k}) - \min_{j}(\widehat{C}_{m,j,k})}{2}$
\State get $\tau_k \gets \mathrm{avg}_i(\tau_{m,k})$
\ForEach {$\hat{j} \in [j]$}
\If {$\widehat{C}_{m,\hat{j},\hat{k}} > 0$ }
\State $\widehat{C}_{m,\hat{j},\hat{k}} \gets  \widehat{C}_{m,\hat{j},\hat{k}} + \min(\tau_{m,\hat{k}}, \tau_{\hat{k}})$
\Else
\State $\widehat{C}_{m,\hat{j},\hat{k}} \gets  \widehat{C}_{m,\hat{j},\hat{k}} - \min(\tau_{m,\hat{k}}, \tau_{\hat{k}})$
\EndIf
\EndFor
\State $I_{\mathrm{perturbed}} \gets \widetilde{M}_{\mathrm{dec}}(\widehat{C})$ 
\end{algorithmic}
\end{algorithm}

Finally, we have two new sets of training samples generated with $a=0$ and $a=1$. We combine these two sets and obtain $50$ random samples per class (user's discretion), to formulate the new dataset $D_{\mathrm{perturbed}}$. We combine $D_{\mathrm{perturbed}}$ and $D_{\mathrm{original}}$, which will effectively increase the number of training samples, solving our target problem. Subsequently, as illustrated by Fig. \ref{fig:sys_5}, we train a new model, $M_2$ with the new $D_{\mathrm{original}}+D_{\mathrm{perturbed}}$ dataset, re-train the decoder with the proposed re-training technique and obtain the final model for character classification.
\subsection{Various Reconstruction Loss Functions} \label{sec:meth_loss}
We investigate the effect on reconstruction based on the loss function used for reconstruction in a capsule network, in order to identify a well-suited reconstruction loss function for the TextCaps model.  First, we study the variations on reconstruction with different loss functions for different number of training samples on the EMNIST-Balanced dataset, and then we extend our analysis to various combinations of loss functions. In this analysis, both spatial and structural similarity measures are used as reconstruction  loss functions. 
 
Since different loss functions we use produce outputs in different scales, it is not possible compare the losses directly. Therefore, we use Peak Signal-to-Noise Ratio (PSNR) given by (\ref{psnr}), as an independent (of reconstruction loss functions) measure, to determine the quality of reconstructed images. 
\vspace{-4mm}
\begin{equation}
PSNR = 10\log_{10} \left(\frac{MAX_i^2}{MSE}\right)
\label{psnr}
\vspace{-1mm}
\end{equation}

where $MAX_i$ is the maximum possible pixel value of the image ($1$ in our case) and $MSE$ is the Mean Squared Error between the test and reconstructed images. 

Let $x(p)$ be the intensity of the $p^{th}$ reconstructed pixel and $y(p)$ be the intensity of the $p^{th}$ true input pixel and $N$ be the total number of pixels.

\vspace{-0.4cm}
\subsubsection{MSE}

Following Sabour \textit{et al.} \cite{sabour2017dynamic}, we use $MSE$, as the loss function for reconstruction, defined by,
\vspace{-0.2cm}
\begin{equation}
MSE = \frac{1}{N} \sum_{i=1}^{N} (x(p)-y(p))^2
\label{mse}
\end{equation}

\vspace{-0.9cm}
\subsubsection{$L_1$ Norm}

To remove artifacts introduced by $MSE$, we consider $L_1$ norm as a loss function which is defined by,
\vspace{-0.3cm}
\begin{equation}
L_1 = \frac{1}{N} \sum_{i=1}^{N} |x(p)-y(p)|
\label{mae}
\end{equation}

\vspace{-0.9cm}
\subsubsection{SSIM}
$L_1$ and $MSE$ do not capture the spatial relationship between pixels. We use $SSIM$ proposed in \cite{Wang:2004:IQA:2319031.2320551} to capture spatial relationship between the input image and reconstructed image. $SSIM$ for $x,y$ and the loss function for $SSIM$, structural dissimilarity ($DSSIM$), are defined by,
\vspace{-0.1cm}
\begin{equation}
SSIM(x,y) = \frac{(2\mu_x\mu_y + C_1)} {(\mu_x^2 + \mu_y^2+C_1)} .\frac{ (2 \sigma _{xy} + C_2)} {(\sigma_x^2 + \sigma_y^2+C_2)}
\label{ssim}
\end{equation}
\vspace{-0.3cm}
\begin{equation}
DSSIM = \frac{1}{N} \sum_{i=1}^{n} 1- SSIM(p)
\label{dssim}
\end{equation}
where $\mu_x$, $\mu_y$ and $\sigma_x^2$, $\sigma_y^2$ are the means and variances and $\sigma _{xy}^2$ is the covariance of reconstructed and true input pixel intensities.
$C_1=(K_1L)^2$ and $C_2=(K_2L)^2$ where $L$ is the dynamic range of the pixel values (typically, $2^m -1$, where $m$ is the number of bits per pixel) and $K_1$, $K_2$ are small constants ($K_1=0.01$ and $K_2=0.03$).

\vspace{-0.3cm}
\subsubsection{Binary Cross-Entropy (BCE)}
$BCE$ is often used as a measure to identify the difference between two distributions, which is defined by,
\vspace{-0.3cm}
\begin{equation}
BCE = - \frac{1}{N} \sum_{i=1}^{n} [y(p)\log(x(p)) + (1-y(p))\log(1-x(p))]
\label{bce}
\end{equation}

\vspace{-0.5cm}
\subsubsection{Combinations of Loss Functions}
To combine two loss functions, rather than linearly combining two loss equations mathematically, we propose a method which combines two reconstructed images together. We slightly modify the CapsNet decoder network using two decoders, one for each loss function and generate two separate reconstruction outputs. Then we compare the absolute values of the difference of each pixel value between the two reconstructed outputs and the test images independently, and assign the pixel value which is closer to the test image to the final reconstructed output. Different loss function combinations we use here for two decoders are $L_1$  \&  $DSSIM$, $L_1$  \&  $BCE$, $MSE$  \&  $DSSIM$, $MSE$  \&  $BCE$ and $BCE$  \&  $DSSIM$.


%


\section{Experiments and Results} \label{sec:results}

%
%
%
%
For each dataset in Table \ref{table:data}, we train TextCaps on 200 training samples from the training set and test using the whole test set. In order to test the performance of the TextCaps architecture, we also evaluate it by training it on full training sets and testing on full test sets.

\begin{table}[!h]
\caption{Five datasets used to evaluate TextCaps}
\label{table:data}
\centering
\footnotesize
\begin{tabular}{|p{2.7cm}|p{0.6cm}|p{1.1cm}|p{0.35cm}|p{0.35cm}|} 
\hline
\multicolumn{1}{|c|}{Dataset} & \multicolumn{1}{c|}{Classes} & Train samp/class & Train size & Test size \\ [1ex] 
\hline
EMNIST-Balanced\cite{2017arXiv170205373C} &  \multicolumn{1}{c|}{47}  &  \multicolumn{1}{c|}{2,400} & \multicolumn{1}{c|}{112,800}  &   \multicolumn{1}{c|}{18,800}  \\ 
\hline
EMNIST-Letters\cite{2017arXiv170205373C} &  \multicolumn{1}{c|}{26}  &  \multicolumn{1}{c|}{4,800} & \multicolumn{1}{c|}{124,800}  &  \multicolumn{1}{c|}{20,800}  \\
\hline
EMNIST-Digits\cite{2017arXiv170205373C} &  \multicolumn{1}{c|}{10}  &  \multicolumn{1}{c|}{24,000} &  \multicolumn{1}{c|}{240,000}  &   \multicolumn{1}{c|}{40,000}  \\ 
\hline
MNIST\cite{lecun1998mnist} &  \multicolumn{1}{c|}{10}  &  \multicolumn{1}{c|}{6,000} & \multicolumn{1}{c|}{60,000}  &   \multicolumn{1}{c|}{10,000}    \\ 
\hline
Fashion MNIST\cite{DBLP:journals/corr/abs-1708-07747} &  \multicolumn{1}{c|}{10}  & \multicolumn{1}{c|}{6,000} & \multicolumn{1}{c|}{60,000}  &  \multicolumn{1}{c|}{10,000}  \\[1ex]
\hline
\end{tabular}
\vspace{-4mm}
\end{table}

\newcolumntype{C}[1]{>{\centering\arraybackslash}m{#1}}

\subsection{Handwritten Character Classification} \label{sec:results_classification}

Table \ref{table:class1} compares our results to the state-of-the-art. We include the results that we obtained with the full training sets, as well as using only 200 training samples per class. In both instances, we have used the full testing sets shown in Table \ref{table:data}, to report the average accuracies. We use a combination of marginal loss and the reconstruction loss for training as proposed in \cite{sabour2017dynamic}, and further, the training procedure followed for every experiment in this paper is similar to \cite{sabour2017dynamic}. For each dataset, we use ensembling to improve our model accuracy, and to avoid over fitting. We use cyclic learning rates for each 30 epochs, giving us 3 ensemble models with 90 epochs \cite{smith2017cyclical}.




\begin{table}[!h]
\caption{Comparison of TextCaps with state-of-the-art results, the mean and the standard deviation from 3 trials are shown}
\vspace{1mm}
\label{table:class1}
\centering
\footnotesize
\begin{tabular}{|C{2.5cm}|C{2.15cm}|C{2.35cm}|}
\hline
\multicolumn{3}{|c|}{\textbf{EMNIST-Letters}}\\
 \hline
Implementation & With full train set &  With 200 samp/class\\ [0.5ex] 
\hline
 \multicolumn{1}{|l|}{Cohen \textit{et al.} \cite{2017arXiv170205373C}} &  85.15\%  & - \\ 
 \multicolumn{1}{|l|}{Wiyatno\textit{et al.}\cite{DBLP:journals/corr/abs-1803-01900}}  & 91.27\% &  - \\
\hline
 \textbf{TextCaps} & \textbf{95.36 $\pm$ 0.30\%} & \textbf{92.79 $\pm$ 0.30\%} \\ 
\hline \hline


\multicolumn{3}{|c|}{\textbf{EMNIST-Balanced}}\\
 \hline
Implementation & With full train set &  With 200 samp/class\\ [0.5ex] 
\hline
 \multicolumn{1}{|l|}{Cohen \textit{et al.} \cite{2017arXiv170205373C}} & 78.02\% & - \\ 
 \multicolumn{1}{|l|}{Dufourq \textit{et al.} \cite{2017arXiv170909161D}}  & 88.3\% & -\\ 
\hline 
 \textbf{TextCaps} & \textbf{90.46 $\pm$ 0.22\%} & 87.82 $\pm$ 0.25\% \\ 
\hline \hline
 
 





\multicolumn{3}{|c|}{\textbf{EMNIST-Digits}}\\
 \hline
Implementation & With full train set &  With 200 samp/class\\ [0.5ex] 
\hline
 \multicolumn{1}{|l|}{Cohen \textit{et al.} \cite{2017arXiv170205373C}}  & 95.90\% & -\\ 
 \multicolumn{1}{|l|}{Dufourq \textit{et al.} \cite{2017arXiv170909161D}}  & 99.3\% & -\\
\hline 
 \textbf{TextCaps} & \textbf{99.79 $\pm$ 0.11\%} & 98.96 $\pm$ 0.22\% \\
\hline \hline

\multicolumn{3}{|c|}{\textbf{MNIST}}\\
\hline
Implementation & With full train set &  With 200 samp/class\\ [0.5ex] 
\hline
\multicolumn{1}{|l|}{Sabour \textit{et al.} \cite{sabour2017dynamic}} & 99.75\% & - \\ 
\multicolumn{1}{|l|}{Cire{\c s}an \textit{et al.} \cite{DBLP:journals/corr/abs-1202-2745}}  &  99.77\% & - \\ 
\multicolumn{1}{|l|}{Wan \textit{et al.} \cite{pmlr-v28-wan13}}  & \textbf{99.79\%} & -\\
\hline 
 \textbf{TextCaps} & 99.71 $\pm$ 0.18\% & 98.68 $\pm$ 0.30\% \\
\hline \hline

\multicolumn{3}{|c|}{\textbf{Fashion MNIST}}\\
\hline
Implementation & With full train set &  With 200 samp/class\\ [0.5ex] 
\hline
 \multicolumn{1}{|l|}{Xiao \textit{et al.} \cite{DBLP:journals/corr/abs-1708-07747}} & 89.7\% & - \\ 
 \multicolumn{1}{|l|}{Bhatnagar \textit{et al.} \cite{8313740}} & 92.54\% & - \\
 \multicolumn{1}{|l|}{Zhong \textit{et al.} \cite{DBLP:journals/corr/abs-1708-04896}}  & \textbf{96.35\%}  & -\\
\hline 
 \textbf{TextCaps}  & 93.71 $\pm$ 0.64\% & 85.36 $\pm$ 0.79\%\\ 
\hline
\end{tabular}
\vspace{-0.3cm}
\end{table}

First, we describe the results we obtained with full training sets and compare with the state-of-the-art. On EMNIST-letters, we significantly outperform the state-of-the-art Wiyatno \textit{et al.} \cite{DBLP:journals/corr/abs-1803-01900} by 4.09\%. An average accuracy of 90.46\% was achieved by our system for the EMNIST-balanced dataset, which outperforms the state-of-the-art Dufourq \textit{et al.} \cite{2017arXiv170909161D} by 2.16\%. For EMNIST-digits dataset, TextCaps was able to surpass the state-of-the-art achieved by Dufourq \textit{et al.} \cite{2017arXiv170909161D} by 0.49\%. For MNIST and Fashion-MNIST, our system produced sub-state-of-the-art accuracy. Yet, our results are on-par.

Subsequently, we describe and compare the results we obtained with only 200 training samples per class. On EMNIST-letters, we exceed the state-of-the-art results by 1.52\%. However for  EMNIST-balanced, EMNIST-digits, MNIST we were able to achieve the state-of-the-art results. Even though our system did not surpass the state-of-the-art performance, we highlight that we were able to achieve a near state-of-the-art performance using only 8-10\% of the training data.

\subsection{Results of the Proposed Image Data Generation Technique}
\label{sec:results_perturb}
\newcommand{\V}[1]{\parbox[c]{0.25cm}{\includegraphics[scale=0.5]{figures/#1}}}
\newcommand{\W}[1]{\parbox[c]{0.3cm}{\includegraphics[scale=0.5]{figures/#1}}}

In this section, we present the results of the decoder re-training technique and the new image data generation by perturbation technique proposed in Section \ref{sec:meth_perturb}. We evaluate the success as well as the limitations of the two techniques by referring to these results. 

\begin{figure}[!h]
\vspace{-2mm}
\centering
\footnotesize
\begin{tabular}{p{2.4cm}cccccccc} 

(a) Test image & \V{final_test/good_6_ori.png} & \V{final_test/good_7_ori.png}& \V{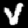}& \V{final_test/good_114_ori.png} & \V{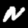} & \V{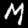} & \V{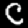} & \V{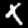}\\
 \cline{1-1}

(b) Recon by $M_{1,\mathrm{dec}}$ & \V{final_ori/good_6_ori.png} & \V{final_ori/good_7_ori.png}& \V{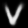}& \V{final_ori/good_114_ori.png} & \V{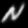} & \V{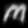} & \V{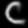} & \V{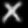}\\
 \cline{1-1}

(c) Recon by $\widetilde{M}_{1,\mathrm{dec}}$ & \V{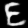} & \V{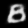}& \V{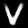}& \V{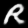} & \V{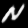} & \V{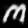} & \V{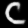} & \V{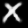}\\


\end{tabular}
\caption{Results of the decoder re-training technique}
\label{fig:unsharp1}
\vspace{-2mm}
\end{figure}

Fig.\ref{fig:unsharp1} (a) illustrates several sample images from the EMNSIT-balanced test set and (b) illustrates the corresponding reconstructed images by $M_{1,dec}$, which was trained using 200 training samples per class. Fig. \ref{fig:unsharp1} (c) illustrates the corresponding reconstructions by $\widetilde{M}_{1,dec}$, which was obtained by re-training $M_{1,dec}$ using the proposed technique. It is evident that reconstructions by $\widetilde{M}_{1,dec}$ are much sharper than those by $M_{1,dec}$. Therefore, our proposed decoder re-training technique is highly successful in sharpening the character reconstruction. 

\begin{figure}[h]
\vspace{-1mm}
\centering
\footnotesize
\begin{tabular}{p{2.4cm}cccccccc} 

(a) Test image & \V{final_test/good_6_ori.png} & \V{final_test/good_7_ori.png}& \V{final_test/good_51_ori.png}& \V{final_test/good_114_ori.png} & \V{final_test/good_75_ori.png} & \V{final_test/good_318_ori.png} & \V{final_test/good_96_ori.png} & \V{final_test/good_261_ori.png}\\
 \cline{1-1}

(b) Recon by $M_{1,\mathrm{dec}}$ & \V{final_ori/good_6_ori.png} & \V{final_ori/good_7_ori.png}& \V{final_ori/good_51_ori.png}& \V{final_ori/good_114_ori.png} & \V{final_ori/good_75_ori.png} & \V{final_ori/good_318_ori.png} & \V{final_ori/good_96_ori.png} & \V{final_ori/good_261_ori.png}\\
 \cline{1-1}


(c) Recon by $\widetilde{M}_{2,\mathrm{dec}}$ & \V{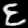} & \V{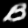}& \V{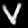}& \V{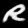} & \V{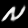} & \V{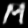} & \V{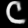} & \V{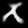}\\

\end{tabular}
\caption{Results of the model $M_2$, trained with the newly generated dataset}
\label{fig:pertub1}
\vspace{-2mm}
\end{figure}

Even though the proposed decoder re-training technique is highly succesful in sharpening the reconstruction, it still does not capture the subtle variations of the input images expected from a succesful reconstruction. Hence, we perform the new image data generation by perturbation technique on the images reconstructed by $\widetilde{M}_{1,\mathrm{dec}}$. Fig. \ref{fig:pertub1} illustrates our results. Fig. \ref{fig:pertub1} (a) and (b) are identical to that of Fig. \ref{fig:unsharp1} (a) and (b), which we include for  comparison. Fig. \ref{fig:pertub1} (c) illustrates the corresponding results obtained by $\widetilde{M}_{2,\mathrm{dec}}$, which was trained afresh using the new dataset generated by the proposed technique. It is evident that reconstructions by $\widetilde{M}_{2,\mathrm{dec}}$ now captures the subtle variations that we required ---curvature of E, slanting of B, asymmetry of X---, and are much closer to the test image than the reconstruction by $M_{1,\mathrm{dec}}$. We observed similar improvements in approximately 80\% of the test images, rendering the proposed technique significantly successful in improving the decoder performance with small number of training samples.

\begin{figure}[!h]
\vspace{-1mm}
\centering
\footnotesize
\begin{tabular}{p{3.5cm}ccc} 
(a) Test image & \W{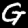} & \W{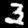}& \W{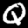} \\
 \cline{1-1}
(b) Recon by $M_{1,\mathrm{dec}}$  & \W{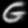} & \W{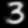}& \W{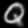} \\
 \cline{1-1}
(c) Recon by $\widetilde{M}_{2,\mathrm{dec}}$  & \W{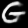} & \W{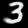}& \W{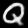} \\
\end{tabular}
\caption{Instances where the model $M_2$ has not been succesful in capturing the subtle variations of the test image ---vertical line of G, bottom part of 3 ---}
\label{fig:pertub2}
\vspace{-2mm}
\end{figure}

Fig. \ref{fig:pertub2} illustrates several instances where the proposed method still failed to capture the required subtle variations. Yet, even in these instances, the reconstructions of $\widetilde{M}_{2,\mathrm{dec}}$ demonstrates significant improvement over the reconstructions of $M_{1,\mathrm{dec}}$.

GANs, VAEs and data augmentation techniques are alternatives to our proposed technique. Fig. \ref{fig:keyswscales} below, illustrates the generated images from a Conditional GAN (CGAN). However, at the generation phase, the new images are generated from random noise and that does not allow to apply specific perturbations, producing less realistic variations in images in comparison to what we do in the proposed method. Similarly, the reconstructions obtained when using our proposed technique are far better than it's alternatives in the low data regime, as shown by Fig. \ref{fig:agumented}, which contains the reconstructions obtained after training with the respective data augmentation technique. The alternatives produced little or reduced improvement, whereas our method produced visually significant ($\approx$ 1dB PSNR) improvement.

\begin{figure}
\centering
\includegraphics[height=6.5mm]{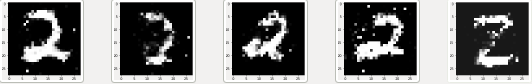}
\caption{Generated images from CGAN for label 2}
\label{fig:keyswscales}
\vspace{-4mm}
\end{figure}

\newcommand{\VI}[1]{\parbox[c]{0.8cm}{\includegraphics[scale=0.6]{images/#1}}}
\begin{figure}[!h]
\begin{tabular}{p{1cm}p{1cm}p{1cm}p{1cm}p{1cm}p{1cm}} 
{\tiny Original} &  {\tiny \hspace{0.2em} Shift(S)} & {\tiny Rotation(R)} & {\tiny \hspace{0.6em} S \& R} & \tiny \hspace{0.4em} Noise & \tiny TextCaps \\
\VI{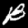} & \VI{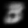}& \VI{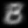} & \VI{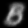} & \VI{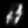} & \VI{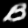}\\
\VI{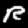} & \VI{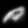}& \VI{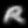} & \VI{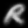} & \VI{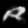} & \VI{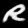} \\
\end{tabular}
\caption{\small Comparison with other data augmentation techniques}

\label{fig:agumented}
\vspace{-4mm}
\end{figure}
\subsection{Results of the Reconstruction Loss Functions and Analysis}
Next, we discuss the variations on reconstruction with respect to different loss functions and combinations of loss functions. All the modifications we consider here were tested with varying number of training samples (100, 200, 500 and 1000) per class, from the \textit{EMNIST-Balanced} dataset. 
We used the CapsNet model proposed in \cite{sabour2017dynamic} with minor alterations for this analysis, where decoder network consists of fully connected layers, since that architecture is well established. 

%

\begin{figure}[!h]
\centering
\subfigure[Change in PSNR for different reconstruction loss functions.\label{psnr_for_loss}]{\includegraphics[scale=0.35]{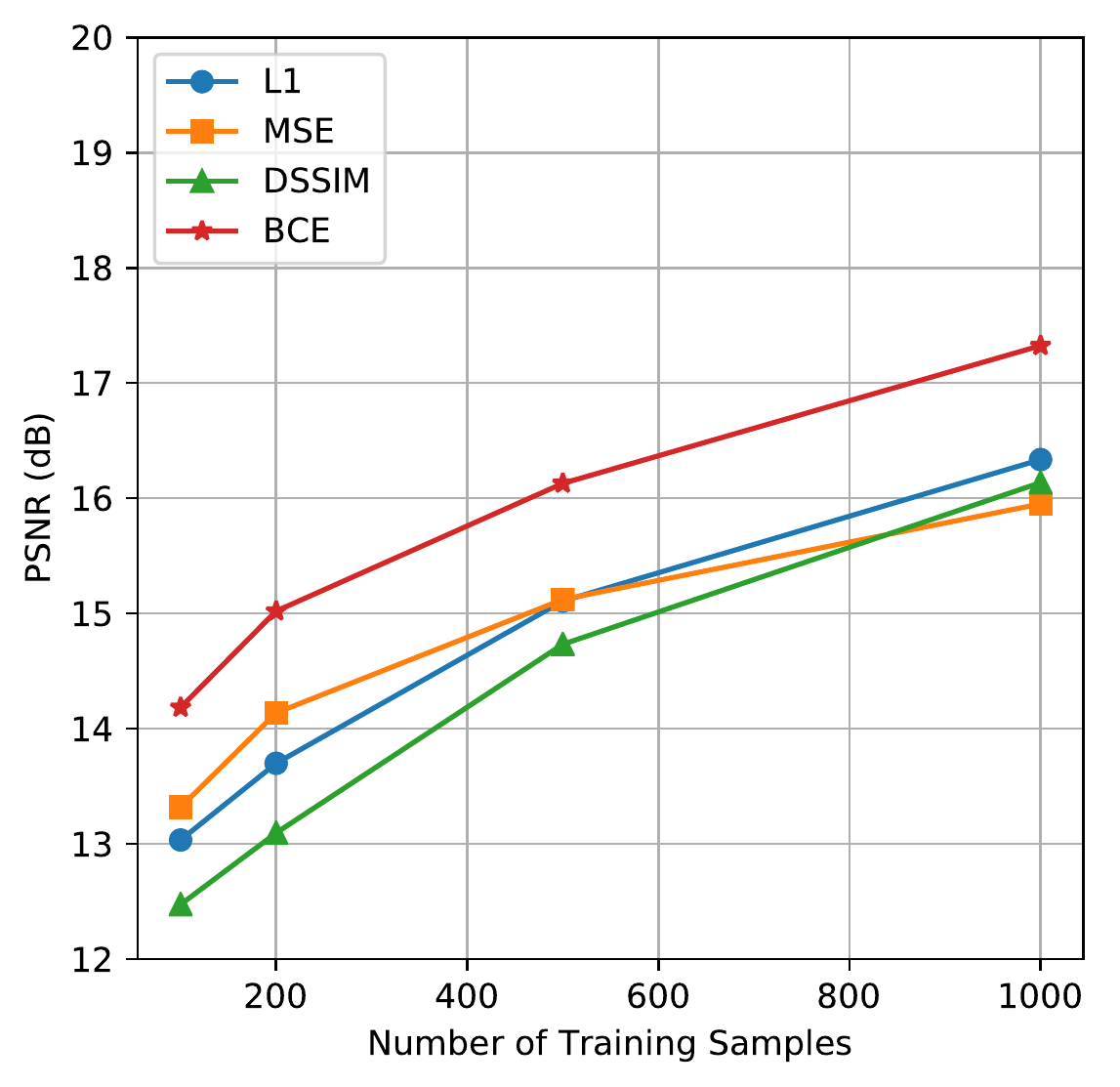}}\hfill
\subfigure[Change in PSNR for different combinations of loss functions.\label{psnr_for_loss_comb}]{\includegraphics[scale=0.35]{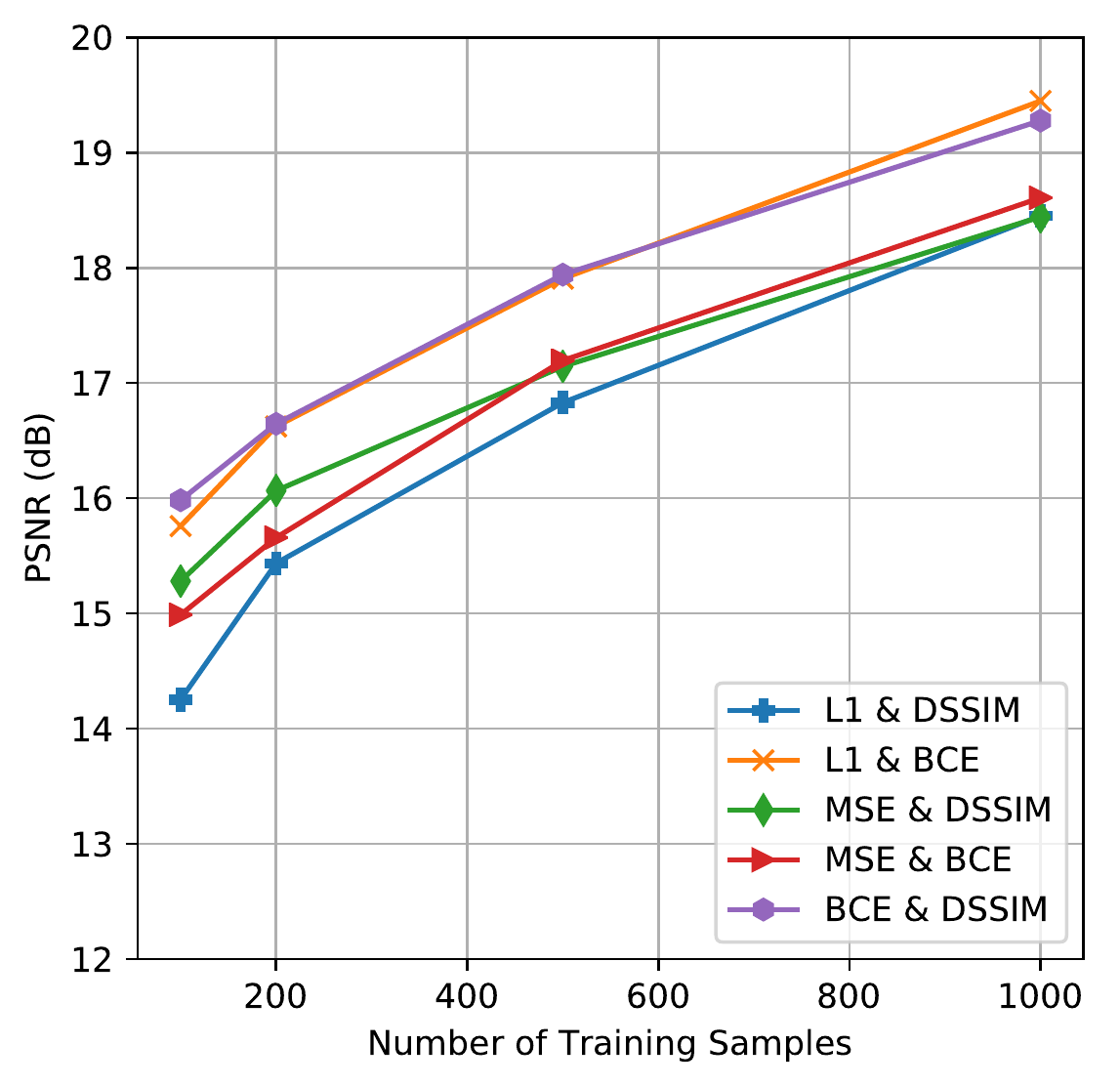}}
\caption{Change in PSNR with the number of training samples.}
\label{psnr_loss_graph}
\vspace{-5mm}
\end{figure}

\begin{figure}[!h]
\vspace{-2mm}
\centering
\footnotesize
\begin{tabular}{m{1.5cm} c }
 Original    	&	\includegraphics[scale=0.325]{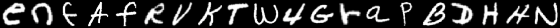}		\\ 
 $L_1$ 			&   \includegraphics[scale=0.325]{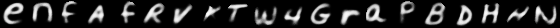}	\\ 
 $MSE$ 			& 	\includegraphics[scale=0.325]{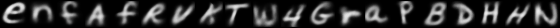}	\\
 $DSSIM$ 		& 	\includegraphics[scale=0.325]{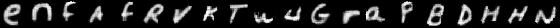}	\\
 $BCE$ 			& 	\includegraphics[scale=0.325]{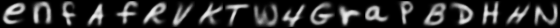}	\\ 
\end{tabular}
\caption{Variations in reconstruction with different loss functions }
\label{real_and_recon_loss}
\vspace{-3mm}
\end{figure}

Fig. \ref{psnr_for_loss} illustrates the variation of PSNR with the amount of training samples used for different loss functions, which leads to a number of interesting observations. For example, when the number of training samples are small (100 or 200), performance of $L_1$ is poor compared to $MSE$, yet for higher number of training samples, $L_1$ performs better. PSNR values for $BCE$ are the highest regardless of the number of training samples. Hence, we conclude that the most suitable loss function for reconstruction loss in general is $BCE$. Fig. \ref{real_and_recon_loss} illustrates the variations in reconstructed images with the use of different loss functions for 200 training samples.

%

We observed that if we linearly combine two loss functions to design a modified loss function, the obtained reconstructions were relatively poor. Nonetheless, since different loss functions capture different properties of handwritten characters, we used two decoders with two loss functions and combined the two reconstructed images together to get a resultant reconstructed image with improved quality. With this modification, it was interestingly observed that individual reconstructions for respective loss functions also improved, due to the effect of the reconstruction loss component in the training loss. However, the amount of improvement of individual PSNRs depends on the loss function combination. Table \ref{table3} demonstrates how the individual PSNR values improve with the two-decoder networks for different combinations of reconstruction loss functions.

\begin{table}[h!]
\caption{PSNR values for individual reconstructions when different combinations of loss functions are used. Here, we use the two-decoder network model with one loss function per each decoder. For each loss function combination, the PSNR value in the first row of PSNR pairs corresponds to the first reconstruction loss function (used in the first decoder) whereas the second row corresponds to the second loss function (used in the second decoder).}
\label{table3}
\centering
\footnotesize
\begin{tabular}{| p{1.9cm} | p{1cm} | p{1cm} | p{1cm} | p{1cm} |} 
 \hline
 \multirow{2}{8em}{Loss function combination} & \multicolumn{4}{c|}{Number of training samples}\\
 \cline{2-5}
 										& 100 		& 200 		& 500 		& 1000 \\ [0.5ex] 
 \hline\hline
 \multirow{2}{4em}{$L_1$  \&  $DSSIM$} 	& 13.51 	& 14.64 	& 15.95		& 17.48  	\\ 
 										& 12.89		& 14.19 	& 15.57		& 17.03	 \\
 \hline
 \multirow{2}{4em}{$L_1$  \&  $BCE$}	& 14.33		& 15.26 	& 16.60		& 18.10  	\\	
 										& 14.57		& 15.44		& 16.71		& 18.13	\\
 \hline
 \multirow{2}{4em}{$MSE$  \&  $DSSIM$}	& 13.87		& 14.81		& 16.00		& 17.29	\\
 										& 12.95		& 14.06		& 15.47		& 16.79 	\\
 \hline												
\multirow{2}{4em}{$MSE$  \&  $BCE$}		& 14.58		& 15.19 	& 16.55		& 17.76  \\
										& 14.59		& 15.20		& 16.56		& 17.78	\\
\hline
 \multirow{2}{4em}{$BCE$  \&  $DSSIM$} 	& 14.62		& 15.41 	& 16.78		& 18.08  \\
 										& 13.80		& 14.77 	& 16.24		& 17.61 \\ [1ex]
 \hline
\end{tabular}
\vspace{-3mm}
\end{table}

\begin{figure}[!h]
\centering
\footnotesize
\begin{tabular}{m{2.2cm} c }
Original         		&	\includegraphics[scale=0.29]{figures/original.png}		\\ 
$L_1$ \& $DSSIM$		&   \includegraphics[scale=0.29]{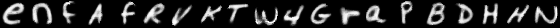}	\\ 
$L_1$ \& $BCE$			& 	\includegraphics[scale=0.29]{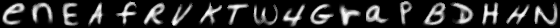}	\\
$MSE$ \& $DSSIM$ 		& 	\includegraphics[scale=0.29]{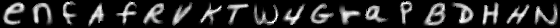}	\\
$MSE$ \& $BCE$			& 	\includegraphics[scale=0.29]{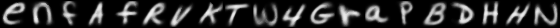}	\\ 
$BCE$ \& $DSSIM$		&	\includegraphics[scale=0.29]{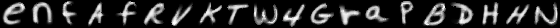}	\\
\end{tabular}
\caption{Variations in reconstruction with different loss function combinations}
\label{real_and_recon_loss_comb}
\end{figure}

With two loss functions, the quality of final reconstructions were much better and PSNR values significantly improved, compared to the single-decoder model with a single loss function. Fig. \ref{psnr_for_loss_comb} shows the improvement in PSNR of the final output reconstruction, when two loss functions are combined together by a two-decoder network. Fig. \ref{real_and_recon_loss_comb} shows the variations in the reconstructed images with the use of different loss function combinations for 200 training samples.

Fig. \ref{psnr_for_loss} illustrates that $BCE$ performs better compared to other loss functions when used in either single-decoder or two-decoder network model. Fig. \ref{psnr_for_loss_comb} illustrates that the combinations $BCE$ \& $DSSIM$ and $L_1$ \& $BCE$ perform significantly better than other loss combinations for the two-decoder model. Even though PSNR values for $DSSIM$ loss were not sufficiently significant, it captures the spatial similarity aspects in reconstruction. Hence, the $BCE$ \& $DSSIM$ loss combination provides marginally better reconstructions, compared to $L_1$ \& $BCE$ for fewer number of training samples. However, when the number of training samples increase, $L_1$ \& $BCE$ combination produces much better reconstructions.

\section{Conclusion} \label{sec:conclusion}
In this paper, we introduced a technique for increasing the size of a dataset by exploiting the concepts in CapsNets. We demonstrated the performance of this technique on well-known handwritten character datasets. Our algorithm takes limited amount of training samples, and perturb their corresponding instantiation parameters to create new training samples. In comparison to the conventional data augmentation techniques in the class of affine transformations, our technique generates images with subtle human-like variations to stroke pattern, boldness and other localized transformations. By combining the original dataset and the perturbed dataset as the training set, we achieved state-of-the-art results and better reconstructions of the input images at the decoder. To further improve the image reconstruction, we analysed the use of different loss functions and their combinations.


Our proposed method works well with images of characters. We intend to extend this framework to images on the RGB space, and with higher resolution, such as images from ImageNet and COCO. Further, we intend to apply this framework on regionally localized languages by extracting training images from font files.
\section{Acknowledgement}
The authors thank the Senate Research Committee of the University of Moratuwa for the financial support through the grant SRC/LT/2016/04 and the Faculty of Information Technology for providing computational resources.

\bibliographystyle{splncs}
\bibliography{references}


\end{document}